\documentclass{article}
\usepackage{ijcai22}

\usepackage{times}

\usepackage{soul}
\usepackage{url}
\usepackage[hidelinks]{hyperref}
\usepackage[utf8]{inputenc}
\usepackage[small]{caption}
\usepackage{graphicx}
\usepackage{amsmath}
\usepackage{booktabs}
\usepackage[nointegrals]{wasysym}
\usepackage{xcolor}
\usepackage{multirow}
\usepackage{amssymb}
\usepackage{subcaption}
\usepackage{stackengine}
\usepackage{float}
\usepackage{graphicx}
\usepackage{booktabs}
\usepackage{times}
\usepackage{amsmath}
\usepackage{wrapfig}
\usepackage{pbox}
\usepackage[export]{adjustbox}
\usepackage[normalem]{ulem}

\title{Triformer: Triangular, Variable-Specific Attentions for Long Sequence Multivariate Time Series Forecasting--Full Version}

\author{
Razvan-Gabriel Cirstea$^1$, 
Chenjuan Guo$^1$, 
Bin Yang$^1$\footnote{Corresponding Author}, 
Tung Kieu$^1$, 
Xuanyi Dong$^2$, 
Shirui Pan$^3$\\
\affiliations
$^1$Aalborg University 
$^2$Google Research, Brain Team 
$^3$Monash University\\
\emails
\{razvan, cguo, byang, tungkvt\}@cs.aau.dk,
xuanyi.dxy@gmail.com,
shirui.pan@monash.edu
}

\begin{document}

\maketitle

\begin{abstract}
A variety of real-world applications
rely on far future information to make decisions, thus calling for \emph{efficient} and \emph{accurate} 
long sequence multivariate time series forecasting. While recent attention-based forecasting models show strong abilities in capturing long-term dependencies, they still suffer from two key limitations. 
First, canonical self attention has a quadratic complexity w.r.t. the input time series length, thus falling short in \emph{efficiency}.  
%
Second, different variables' 
time series often have distinct temporal dynamics, 
which existing studies fail to capture, as they use the same model parameter space, e.g., projection matrices, for all variables' time series, thus 
falling short in \emph{accuracy}.     

To ensure high \emph{efficiency} and \emph{accuracy}, 
we propose \texttt{{Triformer}}, a triangular, variable-specific attention.   
(i) \textit{Linear complexity}: 
%
we introduce a novel patch attention with linear complexity. When stacking multiple layers of the patch attentions, a triangular structure is proposed such that the layer sizes shrink exponentially, thus maintaining linear complexity.  
(ii) \textit{Variable-specific parameters}: we propose a light-weight method to 
enable distinct sets of model parameters for different variables' time series to enhance accuracy without compromising efficiency and memory usage.   
%
Strong empirical evidence on four datasets from multiple domains justifies our design choices, and it demonstrates that \texttt{Triformer} outperforms state-of-the-art methods w.r.t. both accuracy and efficiency.
\end{abstract}

\section{Introduction}

\begin{figure}[t!]
    \centering
    \includegraphics[width=\linewidth]{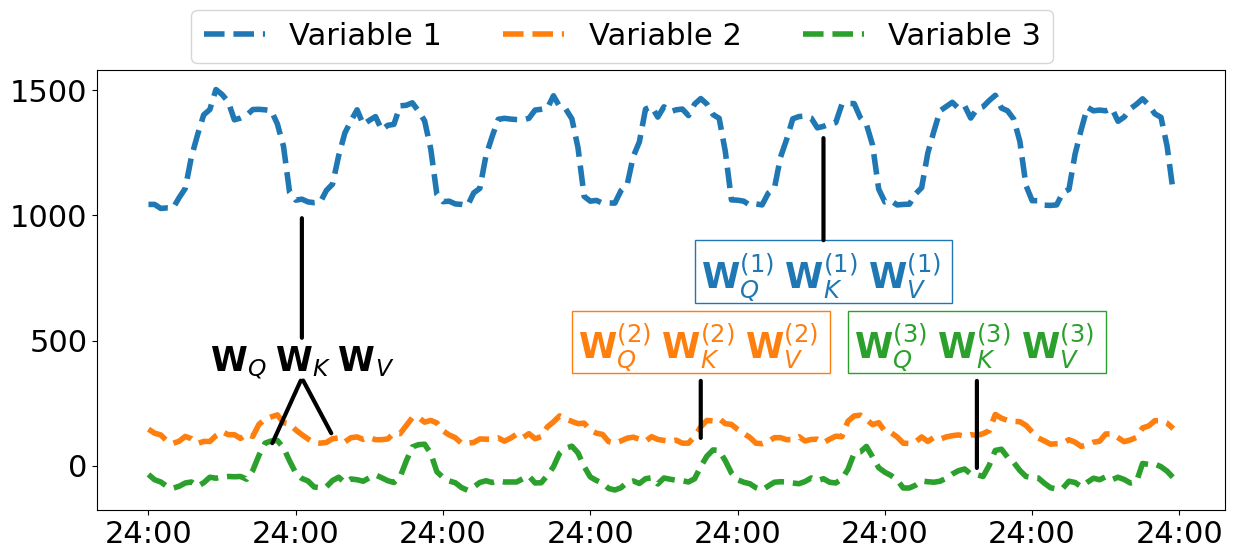}
    \vspace{-1.5em}
    \caption{Example of 3-variate time series, where each variable has a distinct temporal pattern. \textbf{Variable-agnostic modeling:} same projection matrices (see the black matrices) are employed for all variables. \textbf{Variable-specific modeling:} distinct projection matrices (see the colorful matrices) are used for different variables, thus enabling the capture of distinct temporal patterns of different variables.}
    \label{fig:power_example}
\end{figure}

Long sequence multivariate time series forecasting plays an essential role in planning and managing complex systems across diverse domains~\cite{DBLP:conf/icde/LiuJYZ18,DBLP:journals/vldb/GuoYHJC20,DBLP:journals/vldb/PedersenYJ20,DBLP:journals/pvldb/PedersenYJ20}. 
%
In such settings, multiple sensors are often deployed to collect diverse information related to a complex system, thus giving rise to multivariate time series~\cite{MileTS}. Figure~\ref{fig:power_example} shows an example of a 3-variate time series indicating the power consumption of three clients from a power grid system, where each variable has its own time series. 

While extensive studies on short term forecasting exist, e.g., dozens steps ahead forecasting \cite{lstm,qin2017dual,graphwavenet,mtgnn}, a limited number of studies focus on long term forecasting, e.g., hundreds steps ahead. 
Recent studies show that attentions~\cite{attention_all_you_need,wupvldb} are able to capture better long term dependencies, comparing to Recurrent Neural Networks \texttt{RNN}s~\cite{dcrnn,DBLP:conf/icde/Hu0GJX20,tungicde2022} or Temporal Convolutional Networks \texttt{TCN}s~\cite{graphwavenet,davidpvldb,tungicde2022second}. 
However, two main limitations still exist. 

\begin{figure}[h]
        \includegraphics[width=1\linewidth]{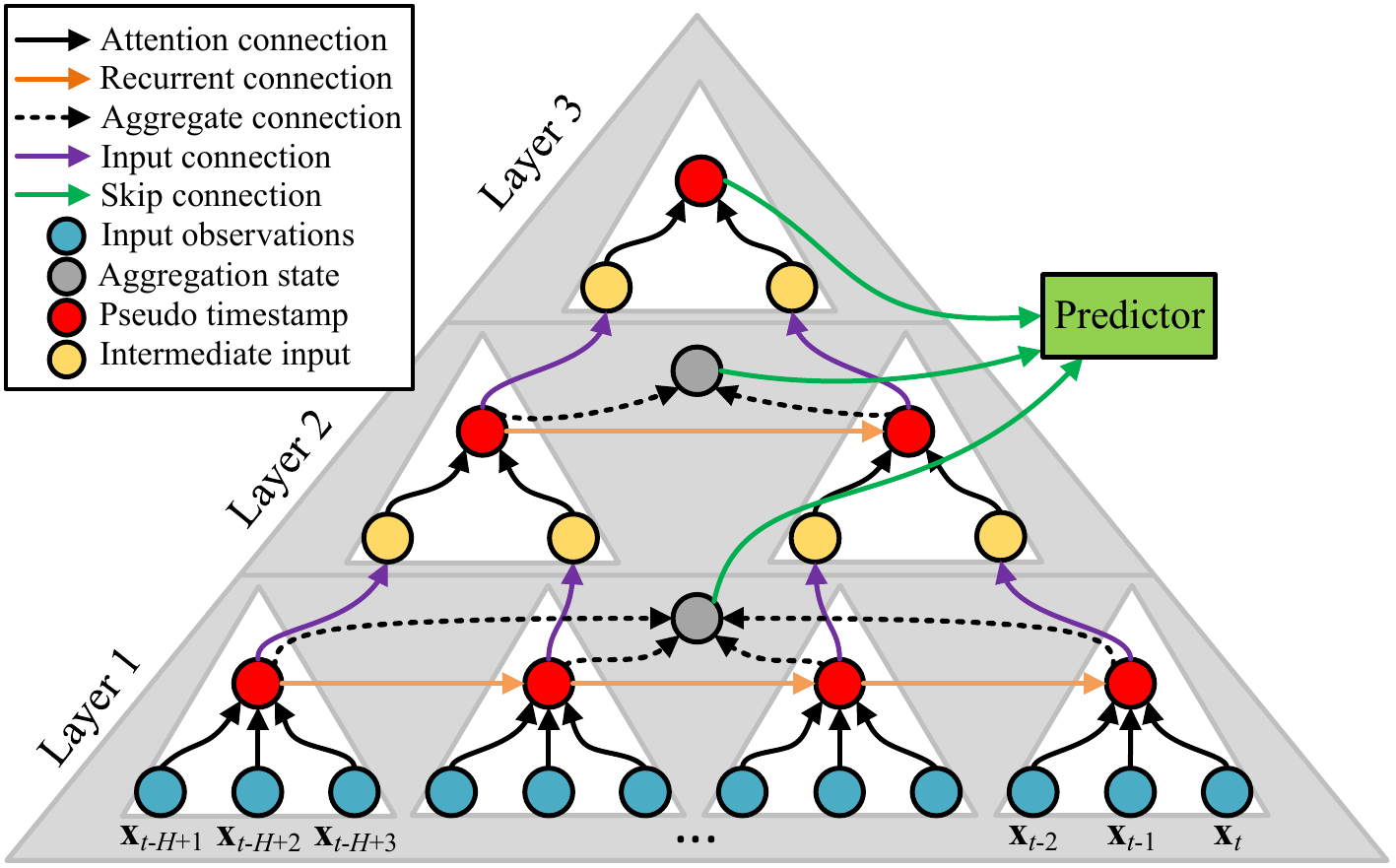}
    \vspace{-1em}
    \caption{\texttt{Triformer} has a triangular structure as the layer input sizes shrink exponentially, making a multi-layer \texttt{Triformer} still with $\mathcal{O}(H)$.}
    \label{fig:attention_example}
\end{figure}

\noindent
\textbf{High complexity: } For a time series of $H$ timestamps, canonical self-attention has a quadratic complexity of $\mathcal{O}(H^2)$. 
Recent studies on long-term forecasting propose different ``sparse'' versions of self-attentions~\cite{informer,reformer}, aiming at reducing the high complexity. 
%
We propose \texttt{Triformer} with linear complexity $\mathcal O(H)$. We first break the time series into small patches. For each patch, we define a novel \emph{Patch Attention} (\texttt{\emph{PA}}) with linear complexity (cf. the small white triangles in Figure~\ref{fig:attention_example}). Specifically, we introduce a pseudo timestamp for a patch and we compute attentions of the timestamps in the patch only to the single pseudo timestamp, making patch attention linear. Then, only the pseudo timestamps are fed into the next layer, such that the layer sizes shrink exponentially, making a triangular structure as shown in Figure~\ref{fig:attention_example}. This ensures a multi-layer \texttt{Triformer} still have linear complexity $\mathcal{O}(H)$.\looseness=-1

\noindent
\textbf{Variable-agnostic parameters: } 
Existing forecasting models often use variable-agnostic parameters, although different variables may exhibit distinct temporal patterns~\cite{graphwavenet,dcrnn,wupvldb}. For example, as shown in Figure~\ref{fig:power_example}, although variable 1 has a unique temporal pattern, which is different from the patterns of variables 2 and 3, the same model parameters, e.g., the projection matrices $\textbf{W}_Q, \textbf{W}_K$, and $\textbf{W}_V$ used in self-attentions (cf. the black matrices in Figure~\ref{fig:power_example}), are applied to all three variables. This forces the learned parameters to capture an ``average'' pattern among all the variables, thus failing to capture distinct temporal pattern of each variable and hurting accuracy.  

We propose a light-weight method to enable variable specific parameters, e.g., a distinct set of matrices $\textbf{W}_Q^{(i)}, \textbf{W}_K^{(i)}$, and $\textbf{W}_V^{(i)}$ for the $i$-th variable (cf. the colorful matrices in Figure~\ref{fig:power_example}), such that it is possible to capture distinct temporal patterns for different variables. Specifically, we factorize the projection matrices into variable-agnostic and variable-specific matrices, where the former are shared among all variables and the latter are specific to different variables. We judiciously make the variable-specific matrices very compact, thus avoiding increasing the parameter space and computation overhead.

We summarize the main contributions as follows: 
(i) We propose a novel, efficient attention mechanism, namely Patch Attention, along with its triangular, multi-layer structure. This ensures an overall linear complexity, thus achieving high \textbf{efficiency}. (ii) We propose a light-weight approach to enable variable-specific model parameters, making it possible to capture distinct temporal patterns from different variables, thus enhancing \textbf{accuracy}. (iii) We conduct extensive experiments on four public, commonly used multivariate time series data sets from different domains, justifying our design choices and demonstrating that the proposal outperforms state-of-the-art methods.
\section{Related Work}\label{section:related_work}
We categorize multivariate time series forecasting methods in 
Table~\ref{table:related_work} along two dimensions---short vs. long term forecasting and variable-agnostic vs. variable-specific modeling.
\begin{table}[h]
\centering
\small
\tabcolsep=0.15cm

\begin{tabular}{|c|c|c|} 
\hline
                                                            & Short Term                                                                                                                                                                               & Long Term                                                                                                                                                                                                                                                      \\ 
\hline
\begin{tabular}[c]{@{}c@{}}Variable\\~Agnostic\end{tabular} & \begin{tabular}[c]{@{}c@{}}\cite{autoformer}\\ \cite{stemgnn}\\\cite{bai2019passenger}\\\cite{graphwavenet}\\\cite{dcrnn}\end{tabular} & \begin{tabular}[c]{@{}c@{}}\cite{informer}\\\cite{reformer}\\\cite{logSparse}\\\cite{stoller2019seq}\\\cite{attention_all_you_need}\end{tabular}  \\ 
\hline
\begin{tabular}[c]{@{}c@{}}Variable\\~Specific\end{tabular} & \begin{tabular}[c]{@{}c@{}}\cite{Razvanicde2022}\\\cite{razvanicde2021}\\\cite{agcrn}\\\cite{KDD_urban_traffic}\end{tabular}                                                                                                  & \texttt{Triformer}                                                                                                                                                                                                                                             \\
\hline
\end{tabular}
\vspace{-0.5em}
\caption{Categorization of Time Series Forecasting.}
\label{table:related_work}
\end{table}

\noindent
\noindent
\textbf{Short vs. Long Term Forecasting}: Most studies focus on short-term forecasting, e.g., 
12 to 48 steps ahead. 
Short-term forecasting models often rely on two types of models that are good at capturing short-term temporal dependencies---recurrent neural networks (\texttt{RNN}s), e.g., \texttt{LSTM} or \texttt{GRU}~\cite{bai2019passenger,dcrnn,lstm_gcn}, and temporal convolutions networks (\texttt{TCN}s), e.g., 1D convolution and \texttt{WaveNet}~\cite{stemgnn,graphwavenet}. 
Both \texttt{RNN}s and \texttt{TCN}s have limited capabilities when dealing with long-range dependencies since they can not directly access the whole input time series, as 
they 
rely on intermediate representations~\cite{khandelwal2018sharp}. Thus, they fall short on long-term forecasting. 
%

For long term forecasting, self-attention based models achieve superior accuracy, but they suffer from quadratic memory and runtime overhead w.r.t. the input time series length $H$. To reduce the complexity, recent long term forecasting studies propose sparse attentions: 
%
\texttt{LogTrans}~\cite{logSparse} is with 
$\mathcal{O}(H(\log H)^2)$, and \texttt{Informer}~\cite{informer} further reduces the complexity to  $\mathcal{O}(H\log H)$. 
There also exist other efficient versions of attentions~\cite{beltagy2020longformer,wang2020linformer,katharopoulos2020transformers}, 
%
%
%
but they are not designed and verified for time series forecasting. 
When stacking multiple layers of attentions, an additional pooling layer is often used after each attention layer. This helps shrink the input size to the next attention layer, thus reducing overall complexity~\cite{dai2020funnel,informer}. 
We propose a novel efficient patch attention with linear complexity, which is able to shrink the layer size by itself without using additional pooling, making a multi-layer patch attention structure also linear.

\noindent
\textbf{Variable-agnostic vs. variable-specific modeling: } Most of the related studies are variable agnostic, meaning that, although different variables' time series may exhibit distinct temporal patterns, they share the same sets of model parameters, e.g., the same weight matrices in \texttt{RNN}s, the same convolution kernels in \texttt{TCN}s, or the same projection matrices in attentions.
Four studies \cite{razvanicde2021,Razvanicde2022,agcrn,KDD_urban_traffic} propose variable-specific modeling for \texttt{RNN}s. Specifically, Cirstea et al. \shortcite{Razvanicde2022,razvanicde2021}
proposed generating variable-specific weight matrices using learned embedding and hyper-networks, however such methods are memory intensive and not suitable for long term forecasting. Pan et al. \shortcite{KDD_urban_traffic} generates variable-specific weight matrices using additional meta-information, e.g., categories of points of interest around the deployed sensor locations. However, such additional meta-information may not be always available. 
Bai et al. \shortcite{agcrn} learns variable-specific weight matrices using a purely data-driven manner without additional information. We empirically compare with Bai et al. \shortcite{agcrn} in the experiments.

\textbf{Key differences from pooling based transformers}: Pooling based transformers such as \cite{dai2020funnel,informer} utilize self-attention as their core component for harvesting temporal dependencies within the data. Self-attention implicitly assumes that each timestamp needs to act as Query, while harvesting information from all the other timestamps which act as Key (thus the name self-attention). Such design implicitly assumes that each Query needs to be dynamic thus conditioned on the input. While such property might be desirable, it has high complexity as each Query needs to attend each Key having quadratic complexity in the case of \cite{dai2020funnel} or $\mathcal{O}(S \; log(S))$ for \cite{informer} (see Figure \ref{fig:selfPatch}). In contrast our proposed Patch-Attention (\texttt{\emph{PA}}) is design on the assumption that a static Query has enough representation power such that, once learned, it can solemnly attend to all Keys (see Figure \ref{fig:PAPatch}). By doing so, not only we reduce the time complexity to linear but we make an attempt into a new direction which does not rely on self-attentions. 

\begin{figure}[h]
    \centering
    \begin{subfigure}[b]{0.5\linewidth}
        \includegraphics[width=1\linewidth]{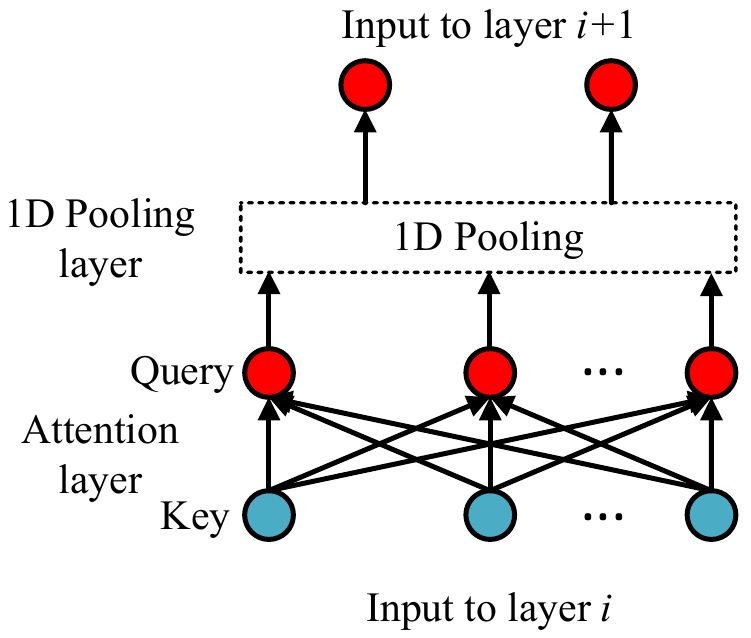}
        \caption{Self-attention, $\mathcal{O}(S^2)$}
        \label{fig:selfPatch}
    \end{subfigure}~~~
    \begin{subfigure}[b]{0.5\linewidth}
        \includegraphics[width=1\linewidth]{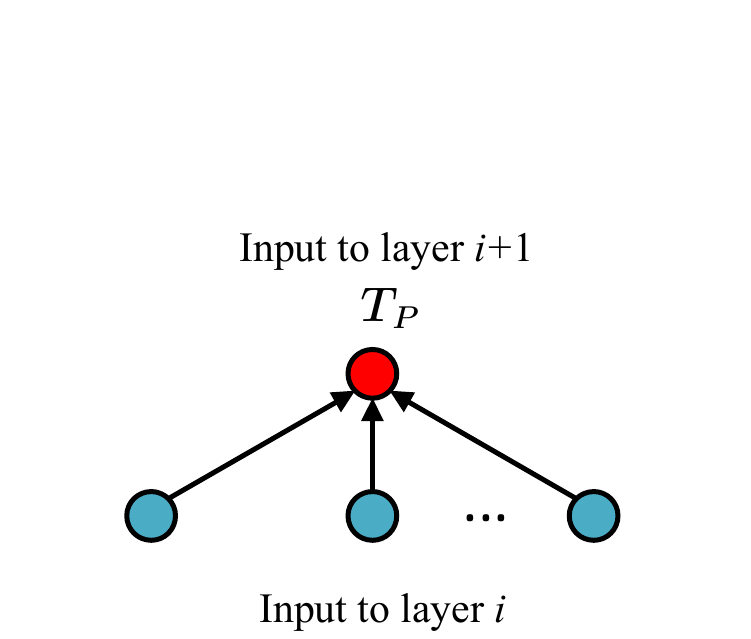}
        \caption{Patch attention, $\mathcal{O}(S)$}
        \label{fig:PAPatch}
    \end{subfigure}
    \caption{Pooling based transformer with self-attention (\texttt{\emph{SA}}) vs. patch attention (\texttt{\emph{PA}}) on patch $p$. In \texttt{\emph{SA}}, each timestamp attends to all other timestamps, thus having $\mathcal{O}(S^2)$. In \texttt{\emph{PA}}, each timestamp only attends to the pseudo timestamp, thus having $\mathcal{O}(S)$. In addition, in \texttt{\emph{SA}}, a pooling layer is applied after each self-attention layer to reduce the input size for the next layer. \texttt{\emph{PA}} does not need pooling anymore, and a learnable Query per patch is fed into the next layer.}
    \label{fig:Forapatch}
\end{figure}

Another important difference is on multi-layer structures. 
To improve model capacity, multiple layers of attentions are stacked together. 
To improve efficiency, it is beneficial to shrink the input sizes of the layers. Our proposal differs from pooling based transformers on the layer size shrinking mechanism. 
In pooling based transformers, a pooling layer is applied after each self-attention layer, as shown in Figure \ref{fig:selfPatch}. 
%
%
The 1D pooling on the temporal axis reduces the size to the next layer.
In contrast, our proposed method does not use any pooling mechanism---for each path, we only allow the learnable Query to move to the next layer after it is updated with information from all Keys. Thus, the size to the next layer is reduced by $1/S$. 

\vspace{1em}
\noindent
\textbf{Key differences from hierarchical transformers}: We design \emph{PA} inspired by hierarchical transformers such as \cite{robert,16x16,pan2021scalable}. While the concept might be similar, there are notable differences between the approaches. First, the previously mentioned studies are design for Nature Language Processing and Computer Vision domains, while our method focuses on time-series. Second the previous mentioned studies try to capture dependencies between different patches, thus resulting in a quadratic complexity w.r.t. the number of patches. In contrast, we propose to introduce a learnable pseudo-timestamp for each patch, such that all timestamps within a specific patch  can write useful information inside.  By doing to we reduce the complexity to linear even after stacking multiple layers. 
\section{Preliminaries} \label{section:pre}

\textbf{Problem Definition. }A multivariate time series records the values of $N$ variables over time.
Observation $\textbf{x}_t \in \mathbb{R}^N$ denotes the values of all $N$ variables at timestamp $t$ and $\textbf{x}^{(i)}_t\in \mathbb{R}$ is the value of the $i$-th variable at $t$. Time series forecasting learns a function $\mathcal{F}$ that takes as into the observations in the historical $H$ timestamps and predicts the future $F$ timestamps. 
    \begin{equation}
    \mathcal{F}_\phi (\mathbf{x}_{t-H+1},..., \mathbf{x}_{t-1},\mathbf{x}_{t}) = (\mathbf{\hat{x}}_{t+1}, \mathbf{\hat{x}}_{t+2},...,\mathbf{\hat{x}}_{t+F} ), 
    \end{equation}
where $\phi$ denotes the learnable parameters of the forecasting model and $\mathbf{\hat{x}}_{j}$ is the predicted values at timestamp $j$. 

\noindent
\textbf{Self-attention. }
Self-attention is a core operation in attention based models. Given a sequence, e.g., a time series, each timestamp attends to all timestamps in the time series, thus being able to derive a representation of the entire time series by capturing both long- and short-term dependencies. %
Consider a time series with $H$ timestamps, e.g., time series $\mathbf{x}^{(i)}\in\mathbb{R}^{H}$ from the $i$-th variable. Canonical self-attention \cite{attention_all_you_need} first transforms the time series into a query matrix $\mathbf{Q}^{(i)}=\mathbf{x}^{(i)}\mathbf{W}_Q\in\mathbb{R}^{H\times d}$, a key matrix $\mathbf{K}^{(i)}=\mathbf{x}^{(i)}\mathbf{W}_K\in\mathbb{R}^{H\times d}$ and a value matrix $\mathbf{V}^{(i)}=\mathbf{x}^{(i)}\mathbf{W}_V\in\mathbb{R}^{H\times d}$, where $d$ represents the hidden representation and $\mathbf{W}_Q, \mathbf{W}_K$, $\mathbf{W}_V \in \mathbb{R}^{d \times d}$ are \textit{projection matrices}, which are learnable.
%
Next, the output is represented as a weighted sum of the values in the value matrix $\mathbf{V}^{(i)}$, where the weights, a.k.a., attention scores, are computed based on the query matrix $\mathbf{Q}^{(i)}$ and the key matrix $\mathbf{K}^{(i)}$, as shown in Equation~\ref{eq:attention}. 

    \begin{equation} \label{eq:attention}
      \mathcal{A}(\mathbf{Q}^{(i)}, \mathbf{K}^{(i)}, \mathbf{V}^{(i)})  =    \varphi
      (\frac{ \mathbf{Q}^{(i)}  \textbf{K}^{(i)T}}{\sqrt{d}} ) \mathbf{V}^{(i)},
    \end{equation}
%
where $\varphi$ represents the \textit{softmax} activation function. 
Computing the attention scores requires quadratic $\mathcal{O}(H^2)$ time complexity and memory usage, which is a major drawback. 
Sparse versions of self-attentions exist,  \cite{logSparse,reformer,informer}, reducing the complexity to $\mathcal{O}(H log(H)^2)$ and $\mathcal{O}(H {\cdot}\log H)$. 
%
We strive for linear complexity.  
In addition, in related attention based works, the same projection matrices $\mathbf{W}_Q$, $\mathbf{W}_K$ and $\mathbf{W}_V$ are applied to all variables $\mathbf{x}^{(i)}$ for $1\leq i\leq N$. In contrast, we strive for variable-specific projection matrices. 

\section{Triformer}

We propose \texttt{Triformer} for learning long-term and multi-scale dependencies in multivariate time series. An overview of the \texttt{Triformer} is shown in Figure~\ref{fig:attention_example}. The design choices of \texttt{Triformer} are three-fold. First, we propose an efficient \emph{Patch Attention} with linear complexity as the basic building block. Second, we propose a triangular structure when stacking multiple layers of patch attentions, such that the layer sizes shrink exponentially. This ensures linear complexity for multi-layer patch attentions and also enables extracting multi-scale features. Third, we propose a light-weight method to enable variable specific modeling, thus being able to capture distinct temporal patterns from different variables, without compromising efficiency. We proceed to cover the specifics of the three design choices.


\subsection{Linear Patch Attention}

To contend with the high complexity, we propose an efficient \emph{Patch Attention} with linear complexity, to ensure competitive overall efficiency. 
Inspired by \cite{pan2021scalable}, we break down the input time series of length $H$ into $P = H/S$ patches along the temporal dimension, where $S$ is the patch size. Figure~\ref{fig:attention_example} shows an example input time series of length $H=12$ being split into $P=4$ patches with patch size $S=3$. We use $\textbf{x}_{p}=\langle \textbf{x}_{(p-1)\cdot S+1}, \ldots, \textbf{x}_{p\cdot S}\rangle$ to denote  the $p$-th patch.

Based on the patches, we compute attention scores per patch. 
If we naively use self-attention in a patch, we still have quadratic complexity (cf. Figure~\ref{fig:selfPatch}). 
To reduce the complexity to linear, for each patch $p$, we introduce a learnable, pseudo timestamp $\textbf{T}_p \in \mathbb{R}^{N \times d}$ (cf. Figure~\ref{fig:PAPatch}). The pseudo timestamp acts as a data placeholder where all the timestamps from the patch can write useful information which is then passed to the next layers.  In \texttt{Triformer}, we choose to use the attention mechanism to update the pseudo timestamp, where the pseudo timestamp works as the Query in self-attentions. The pseudo timestamp queries all the ``real'' timestamps in the patch, thus only computing a single attention score for each real timestamp, giving rise to linear complexity. We call this Patch Attention (\texttt{\emph{PA}}). %

\begin{figure}[h]
    \centering
    \begin{subfigure}[b]{0.5\linewidth}
        \includegraphics[width=1\linewidth]{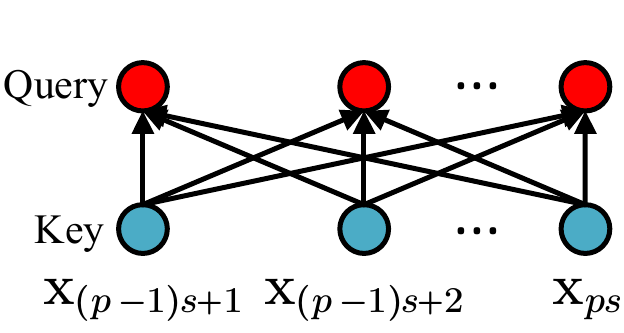}
        \caption{Self-attention, $\mathcal{O}(S^2)$}
        \label{fig:selfPatch}
    \end{subfigure}~~~
    \begin{subfigure}[b]{0.47\linewidth}
        \includegraphics[width=1\linewidth]{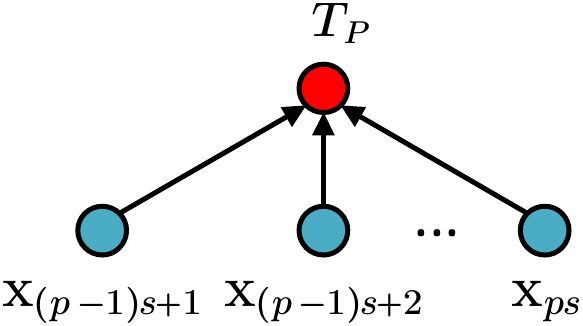}
        \caption{Patch attention, $\mathcal{O}(S)$}
        \label{fig:PAPatch}
    \end{subfigure}
    \vspace{-1.5em}
    \caption{Computing self-attention (\texttt{SA}) vs. patch attention (\texttt{\emph{PA}}) on patch $p$. In \texttt{SA}, each timestamp attends to all other timestamps, thus having $\mathcal{O}(S^2)$. In \texttt{\emph{PA}}, each timestamp only attends to the pseudo timestamp, thus having $\mathcal{O}(S)$.}
    \label{fig:Forapatch}
\end{figure}

Equation~\ref{eq:attention_patch} elaborates the computations of \emph{PA} for pseudo timestamp $\textbf{T}_p$. Note that $\textbf{T}_p=\{\textbf{T}_p^{(i)}\}_{i=1}^N$ and each variable has specific pseudo timestamp $\textbf{T}_p^{(i)}$, thus being variable-specific.

{\small
    \begin{equation} 
    \small
    \label{eq:attention_patch}
      \textbf{T}_p = \mathcal{PA}(\mathbf{T}_p, \mathbf{x}_p)  =  \Bigg\{  \varphi
      (\frac{ \mathbf{T}^{(i)}_p  (\textbf{x}_p^{(i)} \textbf{W}_K)^T}{\sqrt{d}} ) (\textbf{x}_p^{(i)} \textbf{W}_V) \Bigg\}_{i=1}^{N}
    \end{equation}
}

The reduced complexity of \texttt{\emph{PA}} comes with a price that the temporal receptive filed of each timestamp is reduced to the patch size $S$. In contrast, in the canonical attention, it is $H$ covering all timestamps. 
This makes it harder to capture relationships among different patches and also the long-term dependencies, thus adversely affecting accuracy. 
To compensate for the reduced temporal receptive field, we introduce a recurrent connection (cf. the orange arrows in Figure~\ref{fig:attention_example}) to connect the outputs of the patches, i.e., the updated pseudo timestamps according to Equation~\ref{eq:attention_patch},  such that the temporal information flow is maintained. 

As gating mechanism is a crucial component for recurrent networks and has been showed to be powerful controlling the information flow~\cite{information_flow}, we propose a gating rule in the recurrent connections in Equation~\ref{eq:gating}. 

    \begin{equation}
    \label{eq:gating}
    \textbf{T}_{p+1} = g(\boldsymbol{\Theta}_1 \textbf{T}_{p} + \textbf{b}_1) \odot \sigma(\boldsymbol{\Theta}_2 \textbf{T}_{p} + \textbf{b}_2) + \textbf{T}_{p+1},
    \end{equation}
%
where $\boldsymbol{\Theta}_{1}$,  $\boldsymbol{\Theta}_{2}$, $\textbf{b}_{1}$ and $\textbf{b}_{2}$ are learned parameters for the recurrent gates, $\odot$ is  element-wise product, $g(\cdot)$ is \textit{tanh} activation function and $\sigma(\cdot)$ is a \textit{sigmoid} function controlling the information ratio that is passed to the next pseudo timestamp. 

\subsection{Triangular Stacking}

Stacking multiple layers of attention often help improve accuracy. 
In attention based models, each attention layer has the same input size, e.g., the input time series size $H$. In traditional self-attention same input and output have the same shape. Differently pooling-based methods are applying 1D convolution to the output to shrink the temporal horizon. When using \texttt{\emph{PA}}s, we only feed the pseudo timestamps from the patches to the next layer, which shrinks the layer sizes exponentially. More specifically, 
the size of $(l+1)$-th layer is only $\frac{1}{S_l}$ of the size of the $l$-th layer, where $S_l$ is the patch size of the $l$-th layer. This leads to Lemma 1. 
%

\noindent
\textbf{Lemma 1. } \textit{An $L$-layer \texttt{Triformer} has a linear time complexity $\mathcal{O}(H)$ if patch size $S_l \geq 2$ where $1\leq l \leq L$. }

\noindent
\textbf{Proof. } The input size of the $l$-th layer is at most $\frac{H}{\hat{S}^{l-1}}$, where $\hat{S}=\min_{1\leq i \leq L} S_i$ is the minimum patch size of all layers. 
Then, for an $L$-layer \texttt{TRACE}, the sum of the input sizes of all layers is at most $2{\cdot}H$ because 
$$\sum_{i=1}^{L} \frac{H}{\hat{S}^{i-1}} = H \cdot \sum_{i=1}^{L} (\frac{1}{\hat{S}})^{i-1}< \frac{\hat{S}}{\hat{S}-1}\cdot H<2\cdot H,$$ 
where the first inequality holds because $(\frac{1}{\hat{S}})^{i-1}$ forms an exponential series and $\frac{1}{\hat{S}}<1$ and the second inequality hold due to $\hat{S}$ is at least 2. Since the \texttt{\emph{PA}} has a linear complexity w.r.t. the input size. Thus, the complexity of $L$ layers of \texttt{\emph{PA}}s is still linear $\mathcal{O}(H)$.

In a multi-layer \texttt{Triformer}, each layer consists of different numbers of patches and thus having different number of outputs, i.e., pseudo timestamps. 
Instead of only using the last pseudo timestamp per layer, we aggregate all pseudo timestamps per layer into an aggregated output. More specifically, the aggregate output $\textbf{O}^l$ at the $l$-th layer is defined as

    \begin{equation}
        \textbf{O}^l = \theta^l (\textbf{T}_1^l,.., \textbf{T}_k^l,...,\textbf{T}_P^l),
    \end{equation}
%
where $\theta^l$ is a neural network, $\textbf{T}_p^l$ denotes the pseudo timestamp for patch $p\in [1, P]$ at the $l$-th layer. 
%

Finally, the aggregate outputs from all layers are connected to the predictor. This brings two benefits than just using the aggregate output of the last layer. First, the aggregate outputs represent features from different temporal scales, contributing to different temporal views. Second, it provides multiple gradient feedback short-paths, thus easing the learning processes. In the ablation study, we empirically justify this design choice (cf. Table \ref{table:ablation} in Experiments). 

\noindent
\textbf{Predictor. }
We use a 
fully connected neural network as the predictor due to its high efficiency w.r.t. longer term forecasting.\looseness=-1

\subsection{Variable-Specific Modeling}\label{section:variable_modeling}

\begin{figure}[h]
    \centering
    \begin{subfigure}[b]{0.28\linewidth}
        \includegraphics[width=1\linewidth]{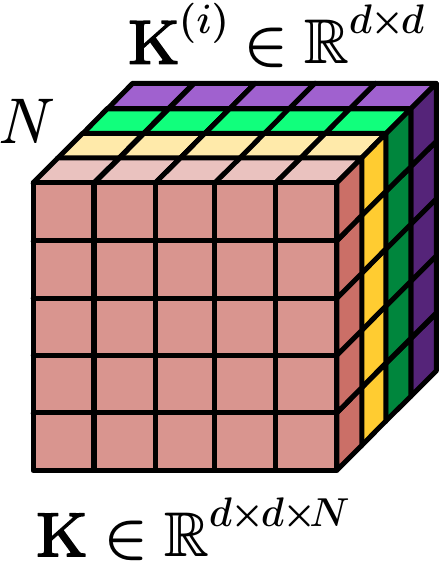}
        \caption{Naive Method}
         \label{fig:projection_naive}
    \end{subfigure}
    \hfill
    \begin{subfigure}[b]{0.6\linewidth}
        \includegraphics[width=1\linewidth]{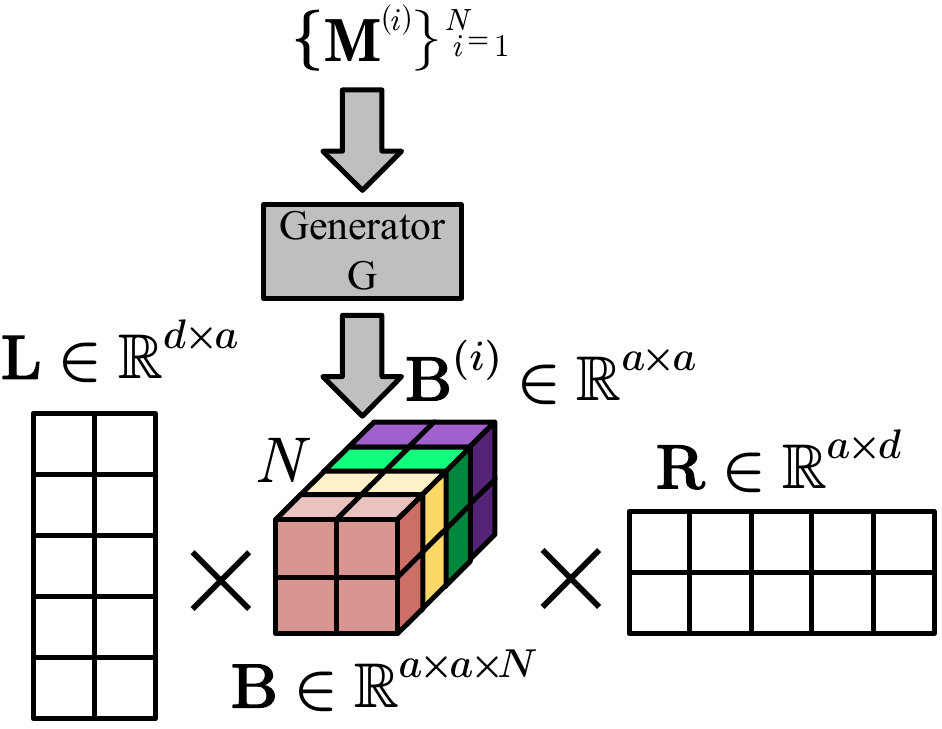}
        \caption{Light-weight Method}
         \label{fig:projection_light}
    \end{subfigure}
    \vspace{-0.3em}
    \caption{Enabling Variable-Specific Projection Matrices. Different colors represent matrices for different variables and $\textbf{L}$ and $\textbf{R}$ are variable-agnostic, thus without colors. For the $i$-th variable, the naive method learns directly matrix $\textbf{K}^{(i)}$, while the light-weight method learns $\textbf{K}^{(i)}$ by $\textbf{L} \times \textbf{B}^{(i)}\times \textbf{R}$, where $\textbf{B}^{(i)}$ is generated by memory $\textbf{M}^{(i)}$ through a generator $\mathcal{G}$.}
    \label{fig:projection}
\end{figure}

Variable-specific modeling can be achieved in a naive way by introducing different projection matrices for each variable, which leads to a very large parameter space. Figure \ref{fig:projection} (a) shows that the naive approach needs to learn $N {\cdot} d^2$ parameters for each projection matrix. 
This may lead to over-fitting, incurs high memory usage, and does not scale well w.r.t. the number of variables $N$.

To contend with the above challenges, we propose a light-weight method to generate variable specific parameters. 
In addition, the method is purely data-driven that only relies on the time series themselves and does not require any additional prior knowledge. 
The overall process is illustrated in Figure \ref{fig:projection} (b). First, we introduce a $m$-dimensional memory vector $\textbf{M}^{(i)} \in \mathbb{R}^m$ for each variable with $i \in [1, N]$. 
The memory is randomly initialized and learnable. This makes the method purely data-driven and can learn the most prominent characteristics of each variable. 

\begin{table*}[h]
\small
\centering
\setlength{\tabcolsep}{0.5em}

\begin{tabular}{|c|c|cc|cc|cc|cc|cc|cc|cc|} 
\hline
\multicolumn{2}{|c|}{Method}                                                                       & \multicolumn{2}{c|}{Reformer} & \multicolumn{2}{c|}{LogTrans} & \multicolumn{2}{c|}{StemGNN}                                                          & \multicolumn{2}{c|}{AGCRN} & \multicolumn{2}{c|}{Informer}                                                         & \multicolumn{2}{c|}{Autoformer}                                                       & \multicolumn{2}{c|}{Triformer}                     \\ 
\hline
                                                                                             & $F$ & MSE   & MAE                   & MSE   & MAE                   & MSE                                       & MAE                                       & MSE   & MAE                & MSE                                       & MAE                                       & MSE                                       & MAE                                       & MSE                     & MAE                      \\ 
\hline
\multirow{5}{*}{$\rotatebox[]{90}{\textbf{ETT}\textsubscript{\textit{h}\textsubscript{1}}}$} & 24  & 0.991 & 0.754                 & 0.686 & 0.604                 & 0.488                                     & 0.508                                     & 0.438 & 0.461              & 0.577                                     & 0.549                                     & \uline{\textcolor[rgb]{1,0.627,0}{0.399}} & \uline{\textcolor[rgb]{1,0.627,0}{0.429}} & \textbf{0.328}          & \textbf{0.380}           \\
                                                                                             & 48  & 1.313 & 0.906                 & 0.766 & 0.757                 & 0.473                                     & 0.505                                     & 0.488 & 0.486              & 0.685                                     & 0.625                                     & \uline{\textcolor[rgb]{1,0.627,0}{0.405}} & \uline{\textcolor[rgb]{1,0.627,0}{0.439}} & \textbf{0.359}          & \textbf{0.401}           \\
                                                                                             & 168 & 1.824 & 1.138                 & 1.002 & 0.846                 & 0.630                                     & 0.577                                     & 0.553 & 0.540              & 0.931                                     & 0.752                                     & \uline{\textcolor[rgb]{1,0.627,0}{0.498}} & \uline{\textcolor[rgb]{1,0.627,0}{0.486}} & \textbf{0.433}          & \textbf{0.449}           \\
                                                                                             & 336 & 2.117 & 1.280                 & 1.362 & 0.952                 & 0.658                                     & 0.598                                     & 0.611 & 0.566              & 1.128                                     & 0.873                                     & \uline{\textcolor[rgb]{1,0.627,0}{0.500}} & \uline{\textcolor[rgb]{1,0.627,0}{0.480}} & \textbf{0.487}          & \textbf{0.475}           \\
                                                                                             & 720 & 2.415 & 1.520                 & 1.397 & 1.291                 & 0.764                                     & 0.675                                     & 0.778 & 0.643              & 1.215                                     & 0.896                                     & \uline{\textcolor[rgb]{1,0.627,0}{0.498}} & \uline{\textcolor[rgb]{1,0.627,0}{0.500}} & \textbf{0.488}          & \textbf{\textbf{0.475}}  \\ 
\hline
\multirow{5}{*}{$\rotatebox[]{90}{\textbf{ETT}\textsubscript{\textit{m}\textsubscript{1}}}$} & 24  & 0.724 & 0.607                 & 0.419 & 0.412                 & \uline{\textcolor[rgb]{1,0.627,0}{0.302}} & 0.381                                     & 0.318 & 0.358              & 0.323                                     & \uline{\textcolor[rgb]{1,0.627,0}{0.369}} & 0.377                                     & 0.417                                     & \textbf{0.252}          & \textbf{0.329}           \\
                                                                                             & 48  & 1.098 & 0.777                 & 0.507 & 0.583                 & \uline{\textcolor[rgb]{1,0.627,0}{0.382}} & \uline{\textcolor[rgb]{1,0.627,0}{0.436}} & 0.426 & 0.429              & 0.494                                     & 0.503                                     & 0.429                                     & 0.442                                     & \textbf{0.275}          & \textbf{0.337}           \\
                                                                                             & 96  & 1.433 & 0.945                 & 0.768 & 0.792                 & \uline{\textcolor[rgb]{1,0.627,0}{0.419}} & \uline{\textcolor[rgb]{1,0.627,0}{0.461}} & 0.409 & 0.433              & 0.678                                     & 0.614                                     & 0.458                                     & 0.460                                     & \textbf{0.314}          & \textbf{0.371}           \\
                                                                                             & 288 & 1.820 & 1.094                 & 1.462 & 1.320                 & \uline{\textcolor[rgb]{1,0.627,0}{0.522}} & \uline{\textcolor[rgb]{1,0.627,0}{0.518}} & 0.545 & 0.531              & 1.056                                     & 0.786                                     & 0.632                                     & 0.526                                     & \textbf{0.385}          & \textbf{0.427}           \\
                                                                                             & 672 & 2.187 & 1.232                 & 1.669 & 1.461                 & 0.644                                     & 0.590                                     & 0.567 & 0.570              & 1.192                                     & 0.926                                     & \uline{\textcolor[rgb]{1,0.627,0}{0.602}} & \uline{\textcolor[rgb]{1,0.627,0}{0.540}} & \textbf{0.437}          & \textbf{0.448}           \\ 
\hline
\multirow{5}{*}{$\rotatebox[]{90}{\textbf{Weather}}$}                                        & 24  & 0.655 & 0.583                 & 0.435 & 0.477                 & 0.377                                     & 0.416                                     & 0.947 & 0.774              & \uline{\textcolor[rgb]{1,0.627,0}{0.335}} & \uline{\textcolor[rgb]{1,0.627,0}{0.381}} & 0.408                                     & 0.445                                     & \textbf{0.323}          & \textbf{0.364}           \\
                                                                                             & 48  & 0.729 & 0.666                 & 0.426 & 0.495                 & 0.438                                     & 0.467                                     & 0.948 & 0.777              & \uline{\textcolor[rgb]{1,0.627,0}{0.395}} & \uline{\textcolor[rgb]{1,0.627,0}{0.459}} & 0.475                                     & 0.487                                     & \textbf{0.390}          & \textbf{0.429}           \\
                                                                                             & 168 & 1.318 & 0.855                 & 0.727 & 0.671                 & \uline{\textcolor[rgb]{1,0.627,0}{0.554}} & \uline{\textcolor[rgb]{1,0.627,0}{0.545}} & 0.950 & 0.774              & 0.608                                     & 0.567                                     & 0.576                                     & 0.550                                     & \textbf{0.497}          & \textbf{0.501}           \\
                                                                                             & 336 & 1.930 & 1.167                 & 0.754 & 0.670                 & 0.598                                     & 0.573                                     & 0.946 & 0.772              & 0.702                                     & 0.620                                     & \uline{\textcolor[rgb]{1,0.627,0}{0.593}} & \uline{\textcolor[rgb]{1,0.627,0}{0.558}} & \textbf{0.538}          & \textbf{0.531}           \\
                                                                                             & 720 & 2.726 & 1.575                 & 0.885 & 0.773                 & \uline{\textcolor[rgb]{1,0.627,0}{0.676}} & \uline{\textcolor[rgb]{1,0.627,0}{0.619}} & 0.938 & 0.770              & 0.831                                     & 0.731                                     & 0.682                                     & 0.634                                     & \textbf{0.587}          & \textbf{0.563}           \\ 
\hline
\multirow{5}{*}{$\rotatebox[]{90}{\textbf{ECL}}$}                                            & 48  & 1.404 & 0.999                 & 0.355 & 0.418                 & 0.209                                     & 0.311                                     & 0.385 & 0.398              & 0.344                                     & 0.393                                     & \uline{\textcolor[rgb]{1,0.627,0}{0.193}} & \uline{\textcolor[rgb]{1,0.627,0}{0.310}} & \textbf{\textbf{0.183}} & \textbf{\textbf{0.279}}  \\
                                                                                             & 168 & 1.515 & 1.069                 & 0.368 & 0.432                 & 0.265                                     & 0.355                                     & 0.481 & 0.509              & 0.368                                     & 0.424                                     & \uline{\textcolor[rgb]{1,0.627,0}{0.223}} & \uline{\textcolor[rgb]{1,0.627,0}{0.334}} & \textbf{0.182}          & \textbf{0.288}           \\
                                                                                             & 336 & 1.601 & 1.104                 & 0.373 & 0.439                 & 0.291                                     & 0.380                                     & 0.564 & 0.561              & 0.381                                     & 0.431                                     & \uline{\textcolor[rgb]{1,0.627,0}{0.232}} & \uline{\textcolor[rgb]{1,0.627,0}{0.345}} & \textbf{0.202}          & \textbf{0.309}           \\
                                                                                             & 720 & 2.009 & 1.170                 & 0.409 & 0.454                 & 0.317                                     & 0.400                                     & 0.725 & 0.678              & 0.406                                     & 0.443                                     & \uline{\textcolor[rgb]{1,0.627,0}{0.257}} & \uline{\textcolor[rgb]{1,0.627,0}{0.361}} & \textbf{0.251}          & \textbf{0.335}           \\
                                                                                             & 960 & 2.141 & 1.387                 & 0.477 & 0.589                 & 0.329                                     & 0.411                                     & 1.005 & 0.829              & 0.460                                     & 0.548                                     & \uline{\textcolor[rgb]{1,0.627,0}{0.265}} & \uline{\textcolor[rgb]{1,0.627,0}{0.364}} & \textbf{0.248}          & \textbf{0.339}           \\
\hline
\end{tabular}

\vspace{-0.5em}
\caption{Overall accuracy. \textbf{Bold} highlights the best results. \uline{\textcolor[rgb]{1,0.647,0}{Underline}} highlights the second best results.}
\vspace{-1.5em}
\label{table:results}

\end{table*}

Next, we propose to factorize a projection matrix, e.g., a Key matrix $\textbf{W}_K^{(i)}\in \mathbb{R}^{d \times d}$ into three matrices---a left variable-agnostic matrix $\textbf{L}_K \in \mathbb{R}^{d \times a}$, a middle variable-specific matrix  $\textbf{B}^{(i)} \in \mathbb{R}^{a \times a}$, and a right variable-agnostic matrix $\textbf{R}_K \in \mathbb{R}^{a \times d}$. We intend to make the middle matrix compact, i.e., $a \ll d$, thus making the method light-weight. 

The left and right matrices are variable-agnostic, thus being shared with all variables. Different variables have their own middle matrices $\{\textbf{B}^{(i)}\}_{i=1}^N$, thus making the middle matrices variable-specific. 
More specifically, the $i$-th variable's middle matrix $\textbf{B}^{(i)}$ is generated from its memory $\textbf{M}^{(i)}$ using a generator $\mathcal{G}(\cdot)$, e.g., a 1-layer neural network. 
This step is also essential to reduce the number of parameters to be learned. 
Learning a full matrix \textbf{B}=$\{\textbf{B}^{(i)}\}_{i=1}^N$ directly requires $N {\cdot} a^2$ parameters. In contrast, when using a generator, it requires $N {\cdot} m$ for the memories, and an additional overhead of $m {\cdot} a^2$ for the generator. 

In addition to the reduced parameter sizes, the factorization also contributes to improved accuracy (cf. Table \ref{table:ablation} in Experiments). Since all variables' time series share the variable-agnostic matrices $\textbf{L}$ and $\textbf{R}$, they act as an implicit regularizer---the amount of combinations that can be produced using them is limited. It also facilitates implicit knowledge sharing among variables.

Equation~\ref{eq:varaible-sp} summarizes the generation of variable-specific Key and Value matrices $\textbf{W}_K^{(i)}$ and $\textbf{W}_V^{(i)}$, which replace $\textbf{W}_K$ and $\textbf{W}_V$ in Equation~\ref{eq:attention_patch} to make Patch Attention variable-specific. 
We do not need the Query matrix as \texttt{\emph{PA}} does not need it and employs the pseudo timestamp instead. 

{\small
    \begin{equation}
    \label{eq:varaible-sp}
    \begin{bmatrix}
    \textbf{W}_K^{(i)} \\ 
    \textbf{W}_V^{(i)} \\ 
    \end{bmatrix}
    =
    \begin{bmatrix}
    \textbf{L}_K \mathcal{G}(\textbf{M}^{(i)}) \textbf{R}_K\\ 
    \textbf{L}_V \mathcal{G}(\textbf{M}^{(i)}) \textbf{R}_V\\
    \end{bmatrix}
    \end{equation}
}

\section{Experiments}

We report on a comprehensive empirical study on four real-world, commonly used time series long term forecasting data sets to justify our design choices and demonstrate that \texttt{Triformer} outperforms the state-of-the-art methods.  

\subsection{Experimental Setup}

\textbf{Datasets:} 
We consider four data sets and follow the setup from the-state-of-the-art method for long term forecasting \cite{informer}.

\noindent
\textbf{ETT}\textsubscript{\textit{h}\textsubscript{1}},  \textbf{ETT}\textsubscript{\textit{m}\textsubscript{1}} (Electricity Transformer Temperature)\footnote{https://github.com/zhouhaoyi/ETDataset}: 
\textbf{ETT}\textsubscript{\textit{h}\textsubscript{1}} and \textbf{ETT}\textsubscript{\textit{m}\textsubscript{1}} has observations every 15 minutes.  Each observation consists of 6 power load features, making them 6-variate time series. To ensure fair comparisons with existing studies, following~\cite{informer}, the train/validation/test data cover 12/4/4 months. 

\noindent
\textbf{ECL} (Electricity Consumption Load)\footnote{https://archive.ics.uci.edu/ml/datasets/ \\ ElectricityLoadDiagrams20112014} is a 321-variate time series, which records the hourly electricity consumption (Kwh) of 321 clients. 
Following~\cite{informer},
the train/validation/test data cover 15/3/4 months.  

\noindent
\textbf{Weather}\footnote{https://ncei.noaa.gov/data/local-climatological-data} is a 12-variate time series which record 12 different climate features, e.g., temperature, humidity, etc, from 1,600 U.S. locations. 
The data is collected hourly. Following~\cite{informer},
the train/validation/test data cover 28/10/10 months. 

\noindent
\textbf{Forecasting Setups:} We use historical $H$ timestamps to forecast the future $F$ timestamps. We vary $H$ and $F$ for different data sets, by following a commonly used long term forecasting setup~\cite{informer}. 
We vary $F$ progressively in \{24, 48, 168, 336, 720, 960\} for the hourly data sets \textbf{ETT}\textsubscript{\textit{h}\textsubscript{1}}, \textbf{ECL}, and \textbf{Weather}, corresponding to 1, 2, 7, 14, 30, and 40 days ahead; and we vary $F$ in 
\{24, 48, 96, 288, 672\} for \textbf{ETT}\textsubscript{\textit{m}\textsubscript{1}} corresponding to 6, 12, 24, 72 and 168 hours ahead. We also vary the input historical timestamps $H$. For \textbf{ETT}\textsubscript{\textit{h}\textsubscript{1}},  \textbf{Weather} and \textbf{ECL}, we vary $H$  from \{24, 48, 96, 168, 336, 720\}, and from \{24, 48, 96, 192, 288, 672\} for \textbf{ETT}\textsubscript{\textit{m}\textsubscript{1}}. Following \cite{informer}, for each $F$ value, we iterate all possible $H$ values and report the best values in Table~\ref{table:results}.

\noindent
\textbf{Baselines:} We select six recent and strong baselines from different categories shown in Table~\ref{table:related_work}, including 
\texttt{StemGNN}~\cite{stemgnn}, \texttt{AGCRN}~\cite{agcrn},  \texttt{Informer}~\cite{informer}, \texttt{Reformer}~\cite{reformer}, \texttt{LogTrans}~\cite{logSparse}, and \texttt{Autoformer}~\cite{autoformer}. 

\noindent
\textbf{Implementation Details:}  We use Adam optimizer with a static learning rate of $1e^{-4}$. We train the models for a maximum of 10 epoch, while utilizing early stopping with a patience of 3 and a batch size of 32. We set $d$=32, $m$=5, and $a$=5 as default values and we further study the effect of different values in the Appendix. Since the patch sizes are highly dependent on the input size $H$, we vary the patch sizes among 2, 3, 4, 6, 7, 12, and 24, and select the best patch sizes based on the validation set. We utilize the same positional embedding mechanism presented in \cite{informer}. All models are trained/tested using a single NVIDIA V100 GPU.

The patch sizes are highly dependent on the input size $H$, we vary the patch sizes $S$ among \{2, 3, 4, 6, 7, 12, 24\} and vary the number of layers $L$ among \{3, 4, 5\}, and select the best patch sizes and the number of layers based on the validation set, which is shown in  in Table \ref{table:patches}. For instance, $(4, 3, 2, 2)$ indicates that we employ a \texttt{Triformer} with 4-layer of patch attentions, where the patch sizes of the 1\textsuperscript{st}, 2\textsuperscript{nd}, 3\textsuperscript{rd}, and 4\textsuperscript{th} layers are 4, 3, 2, and 2, respectively. 
\begin{table}[h]
\centering
\small
\begin{tabular}{|c|c|} 
\hline
$H$ & $S$ and $L$     \\ 
\hline
24                      & (4, 3, 2, 2)        \\ 
\hline
48                      & (4, 3, 4)          \\ 
\hline
96                      & (6, 4, 4)          \\ 
\hline
168                     & (4, 7, 3, 2)        \\ 
\hline
192                     & (6, 4, 4, 2)        \\ 
\hline
288                     & (8, 4, 3, 3)        \\ 
\hline
336                     & (7, 4, 3, 2,~2)  \\ 
\hline
672                     & (7, 6, 4, 4)        \\ 
\hline
720                     & (6, 6, 4)        \\
\hline
\end{tabular}

\caption{Patch Sizes and Number of Layers}
\label{table:patches}
\end{table}

Following \cite{informer}, we consider 3 random training/validation setups, and the results are averaged over 3  runs. For all datasets, we perform standardization such  that the mean of variable is 0 and the standard deviation is 1. 

\subsection{Experimental Results}


Table \ref{table:results} shows the overall accuracy. The results of \texttt{Informer}, \texttt{Reformer} and \texttt{LogTrans} are collected from~\cite{informer}. For \texttt{AGCRN}, \texttt{StemGNN}, and \texttt{Autoformer} we have used their original implementations which are publicly available. \texttt{Triformer} outperforms all baselines on all datasets. In most settings, \texttt{AGCRN} achieves better results when compared to the three attention based methods that are variable-agnostic \texttt{Reformer}, \texttt{LogTrans} and \texttt{Informer}. This highlights the needs of variable-specific modeling. \texttt{Autoformer} achieves the second best accuracy which highlights the importance of auto-correlation. Finally our proposed method, \texttt{Triformer}, achieves the best accuracy in all settings.

\subsection{Experiments for Longer Sequences}

We have performed an addition experiment on the \textbf{ECL} dataset in which we increase $H$ and $F$ to 1024 and 2048 respectively. We have selected the top two most competitive baselines in the form of \texttt{Autoformer} and \texttt{Informer} to compare our method against. The results can be seen in table \ref{table:long_range} where \texttt{Informer} falls behind \texttt{Triformer} while \texttt{Autoformer} runs out of memory (OOM).

\begin{table}[h]
\renewcommand\thetable{1}
\centering
\small
\begin{tabular}{|c|cc|cc|cc|} 
\hline
\begin{tabular}[c]{@{}c@{}}\\\end{tabular} & \multicolumn{2}{c|}{\texttt{Informer}} & \multicolumn{2}{c|}{Autoformer}                & \multicolumn{2}{c|}{\texttt{Triformer}}       \\ 
\hline
H/F                                        & MSE   & MSE                   & MSE                  & MAE                   & MSE            & MAE             \\ 
\hline
1,024                                      & 0.512 & 0.511                 &       OOM               &                 OOM      & \textbf{0.303} & \textbf{0.380}  \\ 
\hline
2,048                                      & 0.941 & 0.804                 & \multicolumn{1}{l}{OOM} & \multicolumn{1}{l|}{OOM} & \textbf{0.350} & \textbf{0.417}  \\
\hline
\end{tabular}
\caption{Long time series, \textbf{ECL}.}
\label{table:long_range}
\end{table}

\noindent
\subsection{Ablation Study}
We perform an ablation study on the \textbf{ECL} dataset, as shown in Table \ref{table:ablation}. First, we remove the variable-specific modeling (VSM), where the same projection matrices are shared across all variables. 
We observe a significant loss of accuracy, which well justifies our design considerations to capture
distinct temporal patterns of different variables. A shared parameter space for all variables fails to do so. 
Second, we replace the light-weight variable specific modeling by the naive way (cf. Figure~\ref{fig:projection}(a)). We observe that the naive method is less accurate and incurs significant more parameters to be learned, which are well aligned with our design considerations.  The generator $\mathcal{G}$ in \texttt{Triformer} takes very limited additional time to generate matrix $\textbf{B}$. 
Third, we do not stack multiple layers patch attentions but use only 1 layer of patch attentions. We observe a significant drop in accuracy, suggesting that the proposed multi-layer, triangular structure, is very effective. 
Fourth, we remove the multi-scale modeling such that only the top most layer's output, rather than the outputs from all layers, is fed into the predictor. 
The accuracy drops significantly, which justifies our design choices of using multi-scale representations in the predictor. 
Finally, we remove the recurrent connections that connect two consecutive pseudo timestamps within \texttt{\emph{PA}}s. The temporal information flow is thus broken, meaning that each patch is computed independently of the others. This results in sub-optimal accuracy. However, we note that the accuracy loss is much smaller than removing other components. 
One of the benefit of this variant is that this makes the computations of all \texttt{\emph{PA}}s per layer fully parallelizable thus improving the running time.

\begin{table}[H]
\vspace{-0.3em}
\centering
\small
\tabcolsep=0.15cm

\begin{tabular}{|c|c|c|c|c|} 
\hline
                                                                             & MSE            & MAE            & \#Param & s/epoch  \\ 
\hline
\texttt{Triformer}                                                                     & 0.183 & 0.279 & 347k    & 73.11    \\ 
\hline
w/o VSM     & 0.215          & 0.273          & 303k    & 33.25    \\ 
\hline
w Naive VSM & 0.191          & 0.288          & 826k    & 63.81    \\ 
\hline
w/o Stacking                                                                 & 0.203          & 0.295          & 258k    & 56.26    \\ 
\hline
w/o Multiscale                                                               & 0.266          & 0.352          & 285k    & 71.58    \\ 
\hline
w/o Recurrent           & 0.191          & 0.290          & 346k    & 65.75    \\
\hline
\end{tabular}
\vspace{-0.5em}
\caption{Ablation Study, \textbf{ECL}.} 
\label{table:ablation}
\vspace{-1.0em}
\end{table}

\noindent
\textbf{Piece-by-Piece Ablation Study}: We conduct a piece-by-piece ablation study (Table \ref{table:ablation}) to study the effects of the three main modules—patch attention (\texttt{\emph{PA}}), triangular stacking (\texttt{\emph{TS}}), and \texttt{\emph{VSM}}. We consider (1) \texttt{PA}, a single \texttt{\emph{PA}} layer; (2) \texttt{PA+TS}, triangular stacking of multiple \texttt{\emph{PA}} layers; (3) \texttt{PA+VSM}, single \texttt{\emph{PA}} layer with \texttt{\emph{VSM}}; and (4) \texttt{Triformer}, i.e., \texttt{\emph{PA}}+\texttt{\emph{TS}}+\texttt{\emph{VSM}}. In addition, to justify the need of recurrent connections (RC) in \texttt{\emph{PA}},
we consider \texttt{PA-RC}, a single layer of \texttt{\emph{PA}} without \texttt{\emph{RC}}. 

\begin{table}[H]
\centering
\small
\setlength{\tabcolsep}{10pt}
\begin{tabular}{|l|c|c|c|c|} 
\hline
             & MSE   & MAE   & \#Param & s/epoch  \\ 
\hline
\texttt{PA} & 0.220 & 0.313 & 244k    & 26.26    \\ 
\hline
\texttt{PA-RC} & 0.223 & 0.314 & 224k    & 22.82 \\ 
\hline
\texttt{PA+TS}  & 0.219 & 0.312 & 309k    & 30.31    \\ 
\hline
\texttt{PA+VSM} & 0.199 & 0.294 & 259k    & 48.92    \\ 
\hline
\texttt{Triformer} & 0.183 & 0.279 & 347k    & 73.11    \\
\hline
\end{tabular}
\caption{Piece-by-Piece Ablation Study, \textbf{ECL}.
}
\label{table:ablation}
\end{table}

Next, we study the effect of the proposed \texttt{VSM} when added to other related studies in the form of \texttt{Informer} \cite{informer}. We can see from Table \ref{table:funnel} that \texttt{VSM} generalize well, significantly increasing the accuracy of \texttt{Informer}.

\begin{table}[h]
\centering
\small
\begin{tabular}{|c|c|c|c|c|c|c|} 
\hline
\begin{tabular}[c]{@{}c@{}}\\\end{tabular} & \multicolumn{2}{c|}{\texttt{Informer}} & \multicolumn{2}{c|}{\texttt{Informer+VSM}} & \multicolumn{2}{c|}{\texttt{Triformer}}     \\ 
\hline
F                                          & MSE   & MAE                   & MAE   & MSE                       & MSE             & MAE              \\ 
\hline
48                                         & 0.344 & 0.393                 & 0.274 & 0.367                     & \textbf{0.183 } & \textbf{0.279 }  \\ 
\hline
168                                        & 0.368 & 0.424                 & 0.284 & 0.376                     & \textbf{0.182 } & \textbf{0.288 }  \\ 
\hline
336                                        & 0.381 & 0.431                 & 0.294 & 0.386                     & \textbf{0.202 } & \textbf{0.309 }  \\ 
\hline
720                                        & 0.406 & 0.443                 & 0.316 & 0.401                     & \textbf{0.251 } & \textbf{0.335 }  \\ 
\hline
960                                        & 0.460 & 0.548                 & 0.317 & 0.396                     & \textbf{0.248 } & \textbf{0.339 }  \\
\hline
\end{tabular}
\caption{Effect of \texttt{VSM}, \textbf{ECL}.}
\label{table:funnel}
\end{table}

\subsection{Hyper-Parameter-Sensitivity Analysis}

We study the impact of the most important hyper-parameters, including $d$, $m$, and $a$. We choose the $\textbf{ECL}$ dataset to perform the experiments since it includes 321-variate (i.e., $N=321$) time series, which has the highest $N$ among all data sets.

In Table~\ref{table:patches}, (4, 4, 3) means that we use patch size 4, 4 and 3 for the 1\textsuperscript{st}, 2\textsuperscript{nd}, and 3\textsuperscript{rd} layers. The first two settings both have 3 layers and we observe similar accuracy, but (2, 6, 4) is slower and with more parameters because the first layer has a smaller patch size and thus more patches. This results in more pseudo timestamps being sent over to the next layer. Next, we observe (12, 2, 2) has a significant accuracy loss compared to the first two 3-layer settings, although the total number of parameters significant drops. When looking at the 2-layer settings, (8, 6) seems to achieve a good trade-off between accuracy and efficiency---it is not as accurate as the first two 3-layer variants but it is more than twice faster. %
To conclude, when using different patch sizes and layers, \texttt{Triformer} offers great flexibility to meet different needs on accuracy, efficiency, and parameter sizes.

\begin{table}[h]
\vspace{-0.3em}
\centering
\small
\begin{tabular}{|l|c|c|c|c|} 
\hline
 & MSE   & MAE   & \#Param & s/epoch  \\ 
\hline
(4, 4, 3)   & 0.183 & 0.279 & 347k    & 73.11     \\ 
\hline
(2, 6, 4)   & 0.181 & 0.276 & 423k    & 99.90      \\ 
\hline
(12, 2, 2)  & 0.381 & 0.436 & 295k    & 37.32     \\ 
\hline
(8, 6)     & 0.213 & 0.305 & 262k    & 35.16     \\ 
\hline
(24, 2)    & 0.801 & 0.743 & 239k    & 27.63     \\
\hline
\end{tabular}
\caption{Sensitivity Analysis of Patch Size $S$, \textbf{ECL}
\vspace{-1.2em}
}
\label{table:patches}
\end{table}

\noindent
\textbf{Impact of $d$}: To study the impact of the size of the hidden representation $d$ used in the pseudo timestamps and projection matrices (cf. subsection ``Linear Patch Attention'' and ``Variable-Specific Modeling''), we vary $d$ among \{16, 32, 64\}. 
The results are shown in Table \ref{table:d_size}. We observe that when $d$ is too small, e.g. $d=16$, the accuracy drops significantly, which indicates that a small $d$ is unable to capture well complex temporal patterns. In addition, the accuracy of $d=64$ is worse than that of $d=32$, which is a possible sign of over-fitting. 

\begin{table}[h!]
\centering
\small
\setlength{\tabcolsep}{20pt}
\begin{tabular}{|c|c|c|} 
\hline
$d$ & MSE & MAE  \\ 
\hline
16   &    0.194    &     0.291       \\ 
\hline
32           & \textbf{0.183}   &  \textbf{0.279}    \\ 
\hline
64     &   0.190  &     0.280 \\ 
\hline
\end{tabular}
\vspace{-0.5em}
\caption{Effect of $d$, \textbf{ECL}}
\label{table:d_size}
\end{table}

\noindent
\textbf{Impact of $m$}: To study the impact of $m$, i.e., the memory size of $\textbf{M}^{(i)}$ (cf. the section ``Variable-Specific Modeling'' and Figure~5 in the main paper), we ran multiple experiments in which we vary $m$ among \{3, 5, 16, 32\}. The results are shown in Table \ref{table:memory_size}. We observe insignificant variations in terms of accuracy, which indicates the proposed method is insensitive w.r.t. the memory size. 

\begin{table}[h!]
\centering
\small
\setlength{\tabcolsep}{20pt}
\begin{tabular}{|c|c|c|} 
\hline
$m$ & MSE & MAE  \\ 
\hline
3           &  0.185    &     0.281 \\ 
\hline
5           & \textbf{0.183}    &    \textbf{0.279}   \\ 
\hline
16          & 0.186    &    0.285        \\ 
\hline
32          &   0.188  &  0.284         \\ 
\hline
\end{tabular}
\vspace{-0.5em}
\caption{Effect of $m$, \textbf{ECL}}
\label{table:memory_size}
\end{table}

\noindent
\textbf{Impact of $a$}: 
We study the impact of $a$, i.e., the size of the middle matrix  $\textbf{B}^{(i)}$ (cf. the section ``Variable-Specific Modeling'' and Figure \ref{fig:projection_light}). We 
vary $a$ among the following values \{3, 5, 16, 32\} and the results are shown in Table \ref{table:a_size}. We observe that when $a$ is too small, e.g. $a=3$, the accuracy drops significantly, which is understandable as the variable-specific matrix $\textbf{B}^{(i)}$ has a shape of $a \times a$, which may be too compact to capture sufficient variable-specific temporal patterns. In addition, we observe that $a=5$ is sufficient and increasing further $a$ to 32 does not contribute to improved accuracy. 

\begin{table}[h!]
\centering
\small
\setlength{\tabcolsep}{20pt}
\begin{tabular}{|c|c|c|} 
\hline
$a$ & MSE & MAE  \\ 
\hline
3           &  0.191    &  0.288    \\ 
\hline
5           & \textbf{0.183}    &    \textbf{0.279}   \\ 
\hline
16     &   0.184  &  0.280   \\ 
\hline
32     &   0.186  &   0.281   \\ 
\hline
\end{tabular}
\caption{Effect of $a$ on \textbf{ECL} Dataset}
\label{table:a_size}
\vspace{-1em}
\end{table}

\begin{figure}[H]
    \vspace{-0.3em}
    \centering
    \includegraphics[width=1\linewidth]{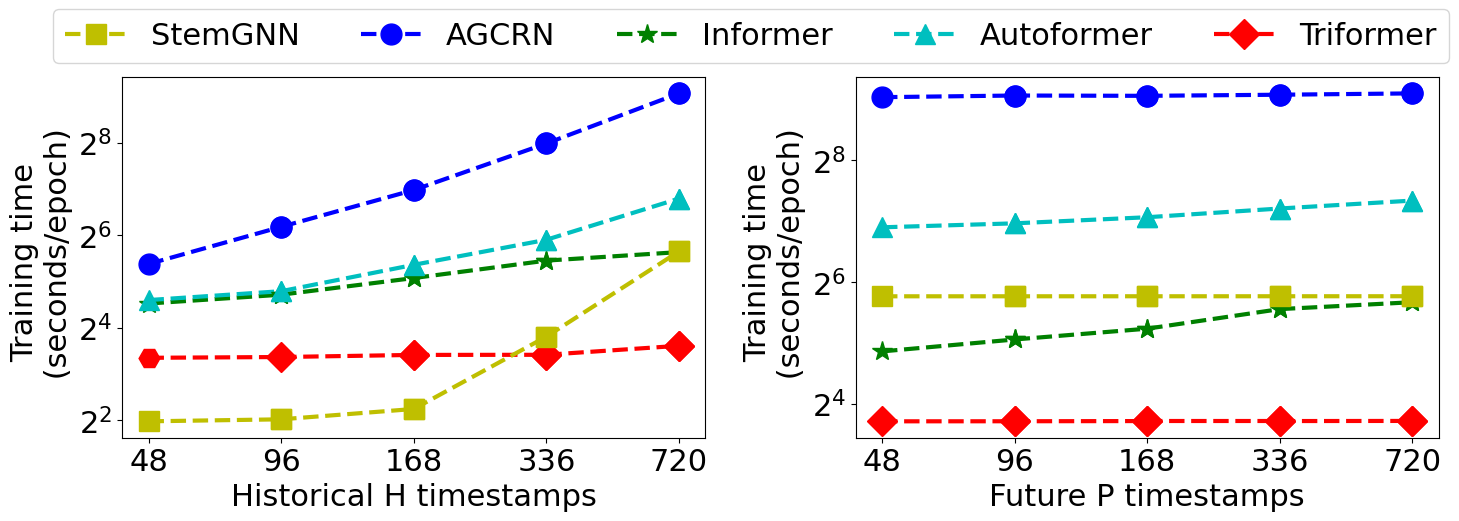}
    \vspace{-1.8em}
    \caption{Training Runtime, seconds/epoch}
    \label{fig:runtime}
    \vspace{-1.0em}
\end{figure}

\noindent
\textbf{Efficiency:} Figure~\ref{fig:runtime} shows the training runtime (seconds/epoch) of \texttt{Triformer} against other four baselines that show the second best accuracy in some data sets, i.e., \texttt{Informer}, \texttt{AGCRN}, \texttt{StemGNN} and \texttt{Autoformer}, as shown in Table \ref{table:results}. 
%
%
When varying $H$: (i) \texttt{Triformer} is faster than \texttt{Informer} and \texttt{Autoformer}, which is in accordance with the complexities of the two methods and also due to the efficient fully connected neural network based predictor. 
(ii) \texttt{StemGNN} increases faster as $H$ increases, while \texttt{Triformer} almost keeps steady. 
(iii) we observe that the recurrent based method \texttt{AGCRN} falls behind when compared with attention based methods, especially when the input time series is long, i.e., large $H$.  
When varying $F$, \texttt{Triformer} is the fastest in all settings. 
In addition, for inference time, all methods are less than 13.3 ms per inference, which are sufficiently efficient to support real-time forecasting. 


\subsubsection{Visualization of Learned Memories.}
\begin{figure}[h]
    \vspace{-1.0em}
    \centering
    \includegraphics[width=\linewidth]{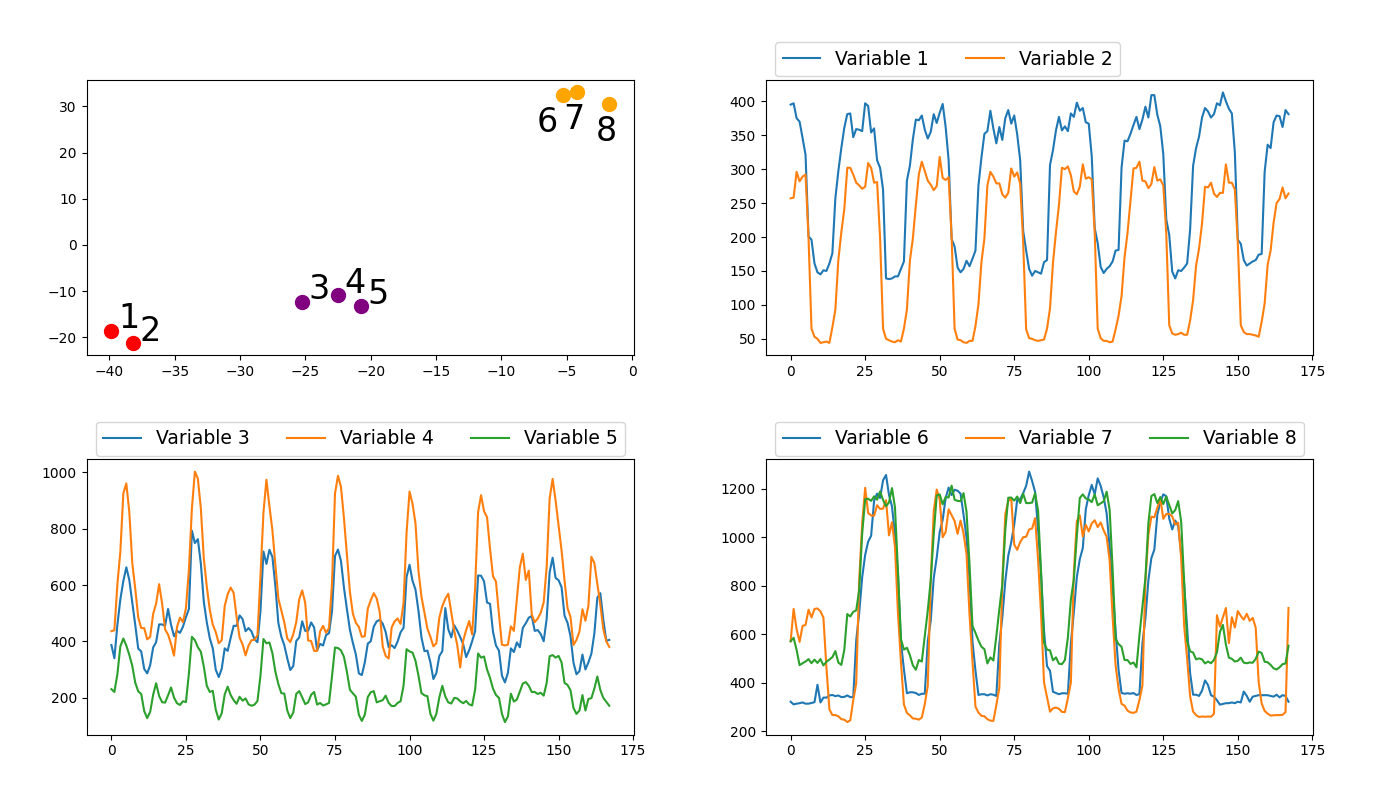}
    \vspace{-2.0em}
    \caption{Visualization of the Learned Memories $\textbf{M}^{(i)}$ for different variables along with their time series.}
    \label{fig:tsne_plot}
    \vspace{-1.0em}
\end{figure}

We visualize the learned memories $\textbf{M}^{(i)}$ to investigate whether they may capture distinct and the most prominent patterns of different variable's time series. 
We select 8 time series from the \textbf{ECL} data sets, as shown in Figure~\ref{fig:tsne_plot}. We use \texttt{t-SNE}~\cite{tsne} to compress each variable memory $\textbf{M}^{(i)}$ to a 2D point, which is also shown in Figure \ref{fig:tsne_plot}. 
We observe that the points are spread over the space, indicating that different time series have their own unique patterns. 
In addition, we observe that when the variables have similar temporal patterns, their learned memories are close by. For example, the time series of variables 6, 7, and 8 are similar, and their corresponding memories are also clustered together, i.e., in the right, top corner. 

\section{Limitations and Possible Solutions} 
A limitation of the proposed Patch Attention \texttt{\emph{PA}} is that the number of learnable parameters is related to forecasting settings. More specifically, a learnable query, i.e., a pseudo-timestamp, is needed per patch. For example, if we train the model on an input size $H=160$ with window size $S=10$, then we learn a total of $16$ pseudo-timestamps. If during testing, we want to change the input size to $H=200$, which requires $40$ pseudo-timestamps, making the already trained model inapplicable.  Thus, this limitation prevents using \texttt{Triformer} in applications that need dynamic input lengths.  

One possible solution is to introduce a generator to generate the learnable pseudo-timestamps from the index of the patch. By doing so, we only need to learn the parameters of the generator network, decoupling the number of patches from the number of learnable pseudo-timestamps, which leads to a more flexible model that addresses the limitation. 

Furthermore, the proposed method cannot forecast different horizons once trained. To remedy this limitation, an encoder-decoder architecture could be used to replace the fully connected predictor, which enables forecasting in an auto-regressive manner.

\section{Conclusion and Outlook}

We propose \texttt{Triformer}, a triangular structure that employs novel patch attentions, which ensures linear complexity. Furthermore, we propose a light-weight method to generate variable-specific projection matrices which are tailored to capture distinct temporal patterns for each variable's time series. Extensive experiments on four data sets show that our proposal outperforms other state-of-the-art methods for long sequence multivariate time series forecasting. In future work, it is of interest to explore different ways of supporting dynamic input lengths and to enhance model training using curriculum learning~\cite{DBLP:conf/ijcai/YangGHT021,SeanIcde2022}. 

\section*{Acknowledgements}
This work was supported in part by Independent Research Fund Denmark under agreements 8022-00246B and 8048-00038B, the VILLUM FONDEN under agreements 34328 and 40567, Huawei Cloud Database Innovation Lab, and the Innovation Fund Denmark centre, DIREC.

\bibliographystyle{named}
\bibliography{references}
\end{document}